\documentclass[11pt]{article}
\pdfoutput=1
\usepackage[utf8]{inputenc}
\usepackage{amssymb,amsmath,amsfonts,geometry,ulem,graphicx,caption,color,setspace,sectsty,comment,footmisc,caption,pdflscape,subcaption,array,hyperref,amsfonts,latexsym,amsthm, mathrsfs, enumerate, mathtools, url,graphicx,float,multirow,listings,xcolor,natbib}

\newtheorem{teo}{Theorem}

\newtheorem{prop}{Proposition}
\newtheorem{defi}{Definition}

\newcolumntype{L}[1]{>{\raggedright\let\newline\\arraybackslash\hspace{0pt}}m{#1}}
\newcolumntype{C}[1]{>{\centering\let\newline\\arraybackslash\hspace{0pt}}m{#1}}
\newcolumntype{R}[1]{>{\raggedleft\let\newline\\arraybackslash\hspace{0pt}}m{#1}}

\begin{document}

\title{Fractional signature: a generalisation of the signature inspired by fractional calculus\thanks{Manuscript published Statistics \& Probability Letters, \url{https://doi.org/10.1016/j.spl.2025.110533}}}
\author{José Manuel Corcuera\thanks{Universitat de Barcelona, Gran Via de les Corts Catalanes, 585, E-08007 Barcelona, Spain. E-mail: \\jmcorcuera@ub.edu. The work of J. M. Corcuera is supported by the Spanish grant PID 118339GB-I00.}, \and Rubén Jiménez\thanks{E-mail: jimenezlum.ruben@gmail.com}}
\date{October 2, 2025}
\maketitle
\begin{abstract}
\noindent In this paper, we propose a novel generalisation of the signature of a path, motivated by fractional calculus, which is able to describe the solutions of linear Caputo controlled FDEs. We also propose another generalisation of the signature, inspired by the previous one, but more convenient to use in machine learning. Finally, we test this last signature in a toy application to the problem of handwritten digit recognition, where significant improvements in accuracy rates are observed compared to those of the original signature.\\
\vspace{0in}\\
\noindent\textbf{MS-Classification 2020:} 60L10, 68Q32, 26A33\\

\bigskip
\end{abstract}

\section{Introduction}
The signature of a path is a sequence of integrals, applied iteratively to the components of the path, which allows it to describe the path precisely and to summarise its characteristics, especially the geometrical ones. This concept emerged in the 1950s and was originally studied by K. T. Chen (for example in \citet{b4}), who gave the first results.

After that, the signature formed part of the Terry Lyons' theory of rough paths (\citet{b9}), which is key in the field of stochastic calculus and in the study of differential equations controlled by rough paths. More recently, the signature has found applications in the field of machine learning (see for example \citet{b5}), where its properties for describing paths are useful for summarising and exploiting data sequences.

From here, we start by recalling some concepts. Let $W = \{(i_{1}, \dots ,i_{k})|k \geq 1,\; i_{1},\dots, i_{k} \in \{1,\dots,d\}\}$ be the set of all multi-indexes given by $d$ values in $\{1, \dots, d\}$, we consider it with the lexicographic order. Then, we can define the signature of a path as follows.

\begin{defi}[Signature of a path]
	Let $X: [a,b]\longrightarrow \mathbb{R}^{d}$ be a path of bounded variation and $I\in W$ a multi-index, $I=(i_{1},\dots,i_{k})$, we define the expressions
	\begin{equation}
		S(X)_{a,b}^{I} \coloneqq \int_{a<t_{1} \dots<t_{k}<b} \quad dX^{i_{1}}_{t_{1}}dX^{i_{2}}_{t_{2}}\dots dX^{i_{k}}_{t_{k}}
	\end{equation} Then, we define the signature of $X$ as $S(X)_{a,b} = \{S(X)_{a,b}^{I}\}_{I\in W}$, where we add 1 as its initial element associated with the multi-index of length 0.
\end{defi}

Recall also the notion of the Riemann-Liouville fractional integral, which coincides with the Caputo one for values of the parameter $0 < \alpha < 1$.

\begin{defi}[Riemann-Liouville fractional integral]
	Let $[a,b] \subseteq \mathbb{R}$ be an interval, let $f:[a,b] \longrightarrow \mathbb{R}$ be a piecewise continuous function and let $\alpha > 0$ be a real parameter, we define the Riemann-Liouville fractional integral of $f$ in the interval $[a,b]$ of order $\alpha$ as the function $I_{{a}^{+}}^{\alpha}f: [a,b] \longrightarrow \mathbb{R}$ given by 
	\begin{equation}
		I_{{a}^{+}}^{\alpha}f(x) \coloneqq \frac{1}{\Gamma(\alpha)}\int_{a}^{x}(x-t)^{\alpha - 1}f(t)dt
	\end{equation}
\end{defi}

\section{Fractional signature}
Once we have recalled these concepts, we can combine them to define a new signature that generalises the original one.
\begin{defi}[Fractional signature of a path]
	Let $X: [a,b]\longrightarrow \mathbb{R}^{d}$ be a $C^{1}$- piecewise path, let  $\alpha > 0 \in \mathbb{R}$ be a parameter and let $I\in W$, $I=(i_{1},\dots,i_{k})$, we can define recursively the expressions ${S}^{\alpha}(X)_{a,b}^{I}$ as follows. If $k = 1$, then: 
	\begin{equation}
		S^{\alpha}(X)_{a,b}^{I} \coloneqq \frac{1}{\Gamma(\alpha)}\int_{a}^{b}(b-t)^{\alpha - 1}dX^{i_{1}}_{t}
	\end{equation}
	
	If $k > 1$, we consider $J = (i_{1},\dots,i_{k-1})$ and then we define: 
	\begin{eqnarray}
	    &S^{\alpha}(X)_{a,b}^{I} \coloneqq \frac{1}{\Gamma(\alpha)}\int_{a}^{b}(b-t)^{\alpha - 1}{S}^{\alpha}(X)_{a,t}^{J}dX^{i_{k}}_{t}
	\end{eqnarray}
	In addition, we then define the fractional signature of $X$ of order $\alpha$ as the sequence ${S}^{\alpha}(X)_{a,b} = \{{S}^{\alpha}(X)_{a,b}^{I}\}_{I\in W}$, where, as before, we add 1 as its initial element associated to the multi-index of length 0.
\end{defi}

First of all, note that we have defined the fractional signature only for $C^{1}$- piecewise paths. That is because their derivative is a piecewise continuous function defined on a closed interval $[a,b]$ and, in particular, is bounded, so the previous integrals are finite and well-defined.

Nevertheless, this is not true in general for bounded variation paths, that constitute the setting where the classical signature is defined (take for example $X: [0,1] \longrightarrow \mathbb{R}$ given by $X_{t} = \sqrt{1-t}$). To avoid focusing on this problem, all paths that we consider will be $C^{1}$- piecewise paths.

From here, it is trivial to see that this fractional signature is a generalisation of the original one, which we can obtain by taking $\alpha = 1$. Then, we can start by easily seeing how it behaves when applied to linear paths. 

\begin{prop}[Fractional signature of a linear path]
	Let $[a,b] \subseteq \mathbb{R}$ be an interval $(a < b)$, let $\alpha > 0$ be a parameter and let $X:[a,b] \longrightarrow \mathbb{R}^{d}$ be a linear path between the points $(A_{1}, \dots, A_{d})$ and $(B_{1}, \dots, B_{d}) \in \mathbb{R}^{d}$, that is, $X_{t}^{i} = \frac{1}{b-a}(A_{i}(b-t) + B_{i}(t-a))$, then for $I \in W$, $I = (i_{1}, \dots, i_{k})$, we have:
	\begin{equation}
		S^{\alpha}(X)_{a,b}^{I} = \frac{\prod_{j = 1}^{k}(B_{i_{j}} - A_{i_{j}})}{\Gamma(\alpha)\Gamma(1 + (k-1)\alpha)}(b-a)^{(\alpha-1)k}\beta((k-1)\alpha + 1, \alpha)
	\end{equation} where $\beta$ denotes the usual beta function.
\end{prop}
\begin{proof}
	Using the fact that $\beta(x, y) = \frac{\Gamma(x)\Gamma(y)}{\Gamma(x+y)}$, it will be enough to show that:
	\begin{equation}
		S^{\alpha}(X)_{a,b}^{I} = \frac{\prod_{j = 1}^{k}(B_{i_{j}} - A_{i_{j}})}{\Gamma(1 + k\alpha)}(b-a)^{(\alpha-1)k} = \frac{\prod_{j = 1}^{k}(B_{i_{j}} - A_{i_{j}})}{\Gamma(1 + k\alpha)(b-a)^{k}}(b-a)^{k\alpha}
	\end{equation}	
	
	Focusing now in this alternative expression and applying the definition of the fractional signature iteratively, it follows that we have that $S^{\alpha}(X)_{a,b}^{I} =$ 
	\begin{eqnarray}
		&\frac{1}{\Gamma(\alpha)^{k}}\int_{a < t_{1} < \dots < t_{k} < b}(b-t_{k})^{\alpha -1}\dots(t_{2}-t_{1})^{\alpha -1}\dot{X}^{i_{1}}_{t_{1}}\dots\dot{X}^{i_{k}}_{t_{k}}dt_{1}\dots dt_{k}\nonumber\\ 
		&= \frac{\prod_{j = 1}^{k}(B_{i_{j}} - A_{i_{j}})}{\Gamma(\alpha)^{k}(b-a)^{k}}\int_{a < t_{1} < \dots < t_{k} < b}(b-t_{k})^{\alpha -1}\dots(t_{2}-t_{1})^{\alpha -1}dt_{1}\dots dt_{k}
	\end{eqnarray} where we have used which is the derivative of the components of the path. So, to end the proof, it will be enough to see, by induction in $k \geq 1$, that: 
	\begin{eqnarray}
		&\int_{a < t_{1} < \dots < t_{k} < b}(b-t_{k})^{\alpha -1}(t_{k}-t_{k-1})^{\alpha -1}\dots(t_{2}-t_{1})^{\alpha -1}dt_{1}\dots dt_{k}\nonumber \\
		& =\frac{\Gamma(\alpha)^{k}}{\Gamma(1+k\alpha)}(b-a)^{k\alpha}
	\end{eqnarray}

	Base step ($k = 1$): it is a direct computation. 
	
	Inductive step: we suppose that the result is true for $k-1 \geq 1$ and now we will see that it also follows for $k$. Indeed, notice that:
	\begin{eqnarray}
		&\int_{a < t_{1} < \dots < t_{k} < b}(b-t_{k})^{\alpha -1}(t_{k}-t_{k-1})^{\alpha -1}\dots(t_{2}-t_{1})^{\alpha -1}dt_{1}\dots dt_{k} = \nonumber\\
		&\int_{a}^{b}(b-t_{k})^{\alpha -1}(\int_{a < t_{1} < \dots < t_{k-1} < t_{k}}(t_{k}-t_{k-1})^{\alpha -1}\dots(t_{2}-t_{1})^{\alpha -1}dt_{1}\dots d_{t_{k-1}})dt_{k}\nonumber\\
		&= \int_{a}^{b}(b-t_{k})^{\alpha -1}\frac{\Gamma(\alpha)^{k-1}}{\Gamma(1+(k-1)\alpha)}(t_{k}-a)^{(k-1)\alpha}dt_{k} = \frac{\Gamma(\alpha)^{k}}{\Gamma(1+k\alpha)}(b-a)^{k\alpha}
	\end{eqnarray} again by a direct computation and the induction hypothesis.
\end{proof}

So, thanks to the above result, we can observe that in the case of linear paths, the fractional signature can be essentially obtained by replacing the monomials $t^{k}$ of the original signature by fractional monomials $t^{\alpha k}$.

From here, it is also easy to check that the fractional signature is invariant by translations of the path, but not by $C^{1}$-time reparametrizations.
\begin{prop}
	Let $X: [a,b]\longrightarrow \mathbb{R}^{d}$ be a path, let $\alpha > 0$ be a parameter and let $c\in \mathbb{R}^{d}$ be a vector, we consider the path $X+c:[a,b]\longrightarrow \mathbb{R}^{d}$ given by $(X+c)_{t} = X_{t}+c$ and then we have that $S^{\alpha}((X+c))_{a,b} = S^{\alpha}(X)_{a,b}$, so the fractional signature of $X$ is invariant by translations.
	
	On the other hand, the fractional signature of order $\alpha \neq 1$ of a path is not invariant by $C^{1}$-time reparametrizations.
\end{prop}
\begin{proof}
	The first part of the statement is trivial because $\dot{(X+c)}^{i}_{t} = \dot{X}^{i}_{t}$ and the fractional signature only depends on the derivatives of the path.
	
	Regarding the second part of the statement, we can give an easy counterexample using linear paths. Let $X:[0,1] \longrightarrow \mathbb{R}$ be the path given by $X_{t} = t$ and let $Y:[0,2] \longrightarrow \mathbb{R}$ given by $Y_{t} = t/2$ be a reparametrization of $X$. Then, if $\alpha \neq 1$, we have that $S^{\alpha}(X)_{0,1}^{1} = \frac{1}{\Gamma(1 + \alpha)}\neq \frac{1}{\Gamma(1 + \alpha)}2^{\alpha - 1} = S^{\alpha}(Y)_{0,2}^{1}$.
\end{proof}

From here on, we can now observe that the fractional signature generalises the property that the classical signature verified in the context of controlled (linear) differential equations. Indeed, we begin by recalling the definition of a Caputo fractional initial value problem and stating a Picard theorem, whose proof can be found or easily derived from the content of the references \citet{b1}, \citet{b8}, \citet{b13}.
\begin{defi}[Solution of a Caputo fractional initial value problem]
	Let $a < h \leq  b \in \mathbb{R}$, let $f:[a,b]\times\mathbb{R}^{n} \longrightarrow \mathbb{R}^{n}$ be a continuous function, let $0 < \alpha \leq 1$ be a parameter and $u_{0} \in \mathbb{R}^{n}$ and let $u: [a, h] \longrightarrow \mathbb{R}^{n}$ be a function, we say that $u$ is a solution of the Caputo fractional initial value problem ${}^{c}D^{\alpha}u = f(t, u)$, $u(a) = u_{0}$ if it verifies for all $t \in [a, h]$ that: 
	\begin{equation}
		u(t) = u_{0} + \frac{1}{\Gamma(\alpha)}\int_{a}^{t}(t-s)^{\alpha -1}f(s, u(s))ds
	\end{equation}
\end{defi}
\begin{teo}[Picard's theorem for linear Caputo fractional differential equations]
	Let $0 < \alpha \leq 1$ be a parameter, $I = [a,b] \subseteq \mathbb{R}$ a closed interval and $\Omega = I \times \mathbb{R}^{n}$. Let also $A:I \longrightarrow \mathbb{L}(\mathbb{R}^{n}, \mathbb{R}^{n})$ be a continuous function, where $\mathbb{L}(\mathbb{R}^{n}, \mathbb{R}^{n})$ can be seen as the space of $n\times n$ real matrices. We then take $f:\Omega \longrightarrow \mathbb{R}^{n}$ given by $f(t,x) = A(t)x$ and it follows that for $(a,u_{0}) \in \Omega$, the Caputo fractional initial value problem given by $f$, $(a,u_{0})$ and $\alpha$ has a unique solution defined in $I$ obtained as the uniform limit of the Picard's iterates $\{u^{n}\}_{n}$ given by $u^{0}(t) = u_{0}$ and 
	\begin{equation}
		u^{n}(t) = u_{0} + \frac{1}{\Gamma(\alpha)}\int_{a}^{t}(t-s)^{\alpha -1}f(s, u^{n-1}(s))ds
	\end{equation}
\end{teo}

Then, we have the following result for the fractional signature.

\begin{teo}
	Let $0 < \alpha \leq 1$, let $X:[a,b]\longrightarrow \mathbb{R}^{d}$ be a $C^{1}$-path and let $V:\mathbb{R}^{e}\longrightarrow \mathbb{L}(\mathbb{R}^{d}, \mathbb{R}^{e})$ be a linear function. Let also $y \in \mathbb{R}^{e}$, then if $Y:[a,b]\longrightarrow \mathbb{R}^{e}$ is a solution of the Caputo fractional initial value problem given by ${}^{c}D^{\alpha}Y = V(Y)\dot{X}_{s},\quad Y_{a} = y$, we have that the value of the Picard's iterate $Y_{t}^{n}$ can be expressed as a linear combination of the elements of the fractional signature $S^{\alpha}(X)_{a,t}$ up to the $n\text{-th}$ level, with coefficients independent of $t$. In particular, $Y_{t}$ is fully determined by $V$, $S^{\alpha}(X)_{a,t}$ and $y$. 
\end{teo}
\begin{proof}
	Indeed, first of all, we note that if we consider $f: [a,b]\times \mathbb{R}^{e}\longrightarrow \mathbb{R}^{e}$ the function given by $f(t,y) = V(y)\dot{X}_{t}$, using the linearity of $V$ it is easy to see that there exists $A:[a,b] \longrightarrow \mathbb{L}(\mathbb{R}^{e}, \mathbb{R}^{e})$ a continuous function such that $f(t,y) = A(t)y$. So the Caputo fractional initial value problem ${}^{c}D^{\alpha}Y = V(Y)\dot{X}_{s},\quad Y_{a} = y$ can be rewritten as ${}^{c}D^{\alpha}Y = f(s, Y),\quad Y_{a} = y$. That expression, combined with the previous theorem, justifies that it has a unique solution and that the Picard's iterates $Y^{n}$ converge to it uniformly. 
	
	Once said that, to show the result we proceed by induction in $n \in \mathbb{N}$.
	
	Base step ($n = 0$): if $n = 0$, then we have that $Y_{t}^{0} = y\cdot 1$, which is what we had to show because the $0\text{-th}$ level of the fractional signature is always 1.
	
	Induction step: assuming that the result is true for $n - 1 \geq 0$, now we will show it for $n$. Indeed, let $t \in [a,b]$, we can see by induction hypothesis that $Y_{s}^{n-1} = \sum_{k = 0}^{n-1}\sum_{i_{1}, \dots,i_{k} \in \{1,\dots, d\}}\lambda_{i_{1},\dots,i_{k}}S^{\alpha}(X)_{a,s}^{i_{1},\dots,i_{k}}$ for some $\lambda_{i_{1},\dots,i_{k}} \in \mathbb{R}^{e}$. Then, we can conclude that: 
	\begin{eqnarray*}
		& Y_{t}^{n} = y + \frac{1}{\Gamma(\alpha)}\int_{a}^{t}(t-s)^ {\alpha - 1}V(Y_{s}^{n-1})\dot{X}_{s}ds = (\text{by the linearity of $V$}) = \nonumber\\
		& y + \sum_{k = 0}^{n-1}\sum_{i_{1}, \dots,i_{k} \in \{1,\dots, d\}}\frac{1}{\Gamma(\alpha)}\int_{a}^{t}(t-s)^{\alpha - 1}V(\lambda_{i_{1},\dots,i_{k}})S^{\alpha}(X)_{a,s}^{i_{1},\dots,i_{k}}\dot{X}_{s}ds = \nonumber
    \end{eqnarray*}
    \begin{eqnarray}
		& y + \sum_{k = 0}^{n-1}\sum_{i_{1}, \dots,i_{k} \in \{1,\dots, d\}}V(\lambda_{i_{1},\dots,i_{k}})\frac{1}{\Gamma(\alpha)}\int_{a}^{t}(t-s)^{\alpha - 1}S^{\alpha}(X)_{a,s}^{i_{1},\dots,i_{k}}\dot{X}_{s}ds = \nonumber\\
		& y + \sum_{k = 0}^{n-1}\sum_{i_{1}, \dots,i_{k} \in \{1,\dots, d\}}V(\lambda_{i_{1},\dots,i_{k}})
		\begin{pmatrix}
			\frac{1}{\Gamma(\alpha)}\int_{a}^{t}(t-s)^{\alpha - 1}S^{\alpha}(X)_{a,s}^{i_{1},\dots,i_{k}}dX_{s}^{1} \\
			\vdots \\
			\frac{1}{\Gamma(\alpha)}\int_{a}^{t}(t-s)^{\alpha - 1}S^{\alpha}(X)_{a,s}^{i_{1},\dots,i_{k}}dX_{s}^{d}
		\end{pmatrix} \nonumber\\
		& = y + \sum_{k = 0}^{n-1}\sum_{i_{1}, \dots,i_{k} \in \{1,\dots, d\}}V(\lambda_{i_{1},\dots,i_{k}})
		\begin{pmatrix}
			S^{\alpha}(X)_{a,t}^{i_{1},\dots,i_{k},1}\\
			\vdots \\
			S^{\alpha}(X)_{a,t}^{i_{1},\dots,i_{k},d}
		\end{pmatrix} = \nonumber\\
		& \sum_{k = 0}^{n}\sum_{i_{1}, \dots,i_{k} \in \{1,\dots, d\}}\mu_{i_{1},\dots,i_{k}}S^{\alpha}(X)_{a,t}^{i_{1},\dots,i_{k}}
	\end{eqnarray} for some $\mu_{i_{1},\dots,i_{k}} \in \mathbb{R}^{e}$ independent of $t$ and $\mu_{0} = y$.
\end{proof}

The above results, as the previous theorem or the proposition that connects the fractional signature with fractional monomials in the linear case, makes us think that this new signature could be more useful in certain settings than the original signature. That is because one can find in the literature many instances where substituting usual calculus, monomials and differential equations by fractional counterparts has proven to be more effective and produce better results, for example in material modelling (\citet{b16}) or regression analysis (\citet{b14}).

\section{Discrete fractional signature}
With that in mind, aiming to show the interest of the fractional signature over the original signature in certain practical applications, we would like to apply and compare both signatures in the field of machine learning. 

Nevertheless, one of the reasons that explains the popularity of the original signature in this field is that it can be easily and quickly computed thanks to the Chen's identity. So, since we have not managed to show a similar identity for the fractional signature, it is not feasible to apply it in machine learning and we cannot make the comparison. 

As a solution to this, we have decided to introduce another generalisation of the original signature, different from the fractional signature and only valid for piecewise linear paths, that verifies a Chen identity and thus can be effectively applied in machine learning. 

\begin{defi}[Discrete fractional signature of a piecewise linear path]
	Let $X: [0,n-1]\longrightarrow \mathbb{R}^{d}$ be a piecewise linear path, that is, there exist points $(A_{i}^{1}, \dots, A_{i}^{d}) \in \mathbb{R}^{d}$ for all $i \in \{0, \dots, n-1\}$ such that: 
	\begin{equation}
		X_{t}^{j} = A^{j}_{i}(i + 1 - t) + A^{j}_{i+1}(t-i) \text{ for all $i \in \{0, \dots, n-2\}$ and }t \in [i, i+1]
		\label{eq:314}
	\end{equation}
	Let $\alpha > 0 \in \mathbb{R}$ be a parameter and let $I\in W$, $I=(i_{1},\dots, i_{k})$, we can define recursively the expressions ${}_{d}DS^{\alpha}(X)_{a,b}^{I}$ for $0 \leq a < b \leq n-1 \in \mathbb{N}$ and $d \in \mathbb{R}$ with $d \geq b$ as follows. If $b = a + 1$, then: 
	\begin{equation}
		{}_{d}DS^{\alpha}(X)_{a,b}^{I} \coloneqq \frac{\prod_{j = 1}^{k}(A^{i_{j}}_{b} - A^{i_{j}}_{a})}{\Gamma(\alpha)\Gamma(1 + (k-1)\alpha)}(\frac{(d-a)^{\alpha}}{b-a})^{k}\beta_{\frac{b-a}{d-a}}((k-1)\alpha + 1, \alpha)
	\end{equation} where $\beta_{z}(x, y) = \int_{0}^{z}t^{x-1}(1-t)^{y-1}dt$ denotes the incomplete beta function and note that the differences $b - a$ of the expression are actually $1$.
	
	If $b > a +1$, then we have that: 
	\begin{eqnarray}
		& {}_{d}DS^{\alpha}(X)_{a,b}^{I} \coloneqq {}_{d}DS^{\alpha}(X)_{a,h}^{I} +  {}_{u}DS^{\alpha}(X)_{a,h}^{i_{1}, \dots, i_{k-1}}\cdot{}_{d}DS^{\alpha}(X)_{h,b}^{i_{k}} \nonumber \\
		& + \dots + {}_{u}DS^{\alpha}(X)_{a,h}^{i_{1}, \dots, i_{r}}\cdot{}_{d}DS^{\alpha}(X)_{h,b}^{i_{r+1}, \dots, i_{k}}+\dots + {}_{d}DS^{\alpha}(X)_{h,b}^{I}
	\end{eqnarray} with $u = \frac{b + h}{2}$ and $h$ the smallest integer greater than or equal to $\frac{a + b}{2}$. 
	
	In addition, then we can define the discrete fractional signature of order $\alpha$ of $X$ as the sequence $DS^{\alpha}(X)_{0,n-1} = \{{}_{n-1}DS^{\alpha}(X)_{0,n-1}^{I}\}_{I\in W}$, where as before we add 1 as its initial element associated to the multi-index of length 0.
\end{defi}

Note that, in this case, the discrete fractional signature already verifies by definition a result analogous to Chen's identity, which was our objective and which will allow us to apply this signature in the field of machine learning. In what follows, we will understand by piecewise linear path an application such as the one described in the previous definition. 

With this definition, we can easily observe that for piecewise linear paths the discrete fractional signature is a generalisation of the classical one.

\begin{prop}
	Let $X: [0,n-1]\longrightarrow \mathbb{R}^{d}$ be piecewise linear path, then we have that $DS^{1}(X)_{0,n-1} = S(X)_{0,n-1}$, that is, for all $I\in W$, $I=(i_{1},\dots, i_{k})$, it is true that ${}_{n-1}DS^{1}(X)_{0,n-1}^{I} = S(X)_{0,n-1}^{I}$. So, for piecewise linear paths, the discrete fractional signature is a generalisation of the signature.
\end{prop}
\begin{proof}
	Indeed, as $X: [0,n-1]\longrightarrow \mathbb{R}^{d}$ is a piecewise linear path, then there exist $(A_{i}^{1}, \dots, A_{i}^{d}) \in \mathbb{R}^{d}$ for all $i \in \{0, \dots, n-1\}$ such that \ref{eq:314} is verified. Then, by definition, the elements of $DS^{1}(X)_{0,n-1} = \{{}_{n-1}DS^{1}(X)_{0,n-1}^{I}\}_{I\in W}$ are given by a recurrence with base case (if $b = a + 1$):
	\begin{eqnarray}
		&{}_{d}DS^{1}(X)_{a,b}^{I} = \frac{\prod_{j = 1}^{k}(A^{i_{j}}_{b} - A^{i_{j}}_{a})}{\Gamma(k)}(\frac{d-a}{b-a})^{k}\beta_{\frac{b-a}{d-a}}(k, 1)\nonumber \\ 
		&= \frac{\prod_{j = 1}^{k}(A^{i_{j}}_{b} - A^{i_{j}}_{a})}{(k-1)!}(\frac{d-a}{b-a})^{k}\frac{(b-a)^{k}}{k(d-a)^{k}}= \frac{\prod_{j = 1}^{k}(A^{i_{j}}_{b} - A^{i_{j}}_{a})}{k!} = S(X)_{a,b}^{I}
	\end{eqnarray}
	Notice that the above does not depend on $d$. Consequently, since in order to calculate a term of this signature we are always reduced to this base case, we can ignore the value appearing in the bottom left of the notation, even when we consider the formula for $b > a + 1$. 
	
	We then observe that the formula for $b > a + 1$ becomes the original Chen's identity. That proves that both signatures are the same.
\end{proof}

From here, although the fractional signature and the discrete fractional signature are two different generalisations of the classical signature, they are related and coincide in certain cases, as shown in what follows.
\begin{prop}
	Let $X: [0,n-1]\longrightarrow \mathbb{R}^{d}$ be a piecewise linear path and let $\alpha > 0$ be a parameter, then we have that ${}_{n-1}DS^{\alpha}(X)_{0,n-1}^{j} = S^{\alpha}(X)_{0,n-1}^{j}$, that is, the first level of both signatures coincide.
	
	In addition, let $Y: [0,1]\longrightarrow \mathbb{R}^{d}$ be a linear path between the points $A_{0}$, $A_{1} \in \mathbb{R}^{d}$, that is, $Y_{t}^{i} = A_{0}^{i}(1-t) + A_{1}^{i}t$, then $DS^{\alpha}(Y)_{0,1} = S^{\alpha}(Y)_{0,1}$.
\end{prop}
\begin{proof}
	Let $j \in \{1,\dots, d\}$ and recall that $X: [0,n-1]\longrightarrow \mathbb{R}^{d}$ is a piecewise linear path, so there exist points $(A_{i}^{1}, \dots, A_{i}^{d}) \in \mathbb{R}^{d}$ for all $i \in \{0, \dots, n-1\}$ such that \ref{eq:314}, then we can see that taking $h$ appropriately we have:
	
	\begin{eqnarray}
		&{}_{n-1}DS^{\alpha}(X)_{0,n-1}^{j} = {}_{n-1}DS^{\alpha}(X)_{0,h}^{j} + {}_{n-1}DS^{\alpha}(X)_{h,n-1}^{j} =\text{ (iteratively) } \nonumber \\
		&=\sum_{i=0}^{n-2}{}_{n-1}DS^{\alpha}(X)_{i,i+1}^{j}= \sum_{i=0}^{n-2}\frac{(A^{j}_{i+1} - A^{j}_{i})}{\Gamma(\alpha)}(n-1-i)^{\alpha}\beta_{\frac{1}{n-1-i}}(1, \alpha) \nonumber\\
		& = \frac{1}{\Gamma(\alpha + 1)}\sum_{i=0}^{n-2}(A^{j}_{i+1} - A^{j}_{i})((n-1-i)^{\alpha} - (n-2-i)^{\alpha}).
	\end{eqnarray} where we have computed the integral of the definition of $\beta_{\frac{1}{n-1-i}}(1, \alpha)$.
	
	Then, to proof the first statement we only need to notice that: 
	\begin{eqnarray}
		& S^{\alpha}(X)_{0,n-1}^{j} = \frac{1}{\Gamma(\alpha)}\int_{0}^{n-1}(n-1-t)^{\alpha - 1}\dot{X}^{j}_{t}dt \nonumber \\
		&=\frac{1}{\Gamma(\alpha)}\sum_{i=0}^{n-2}\int_{i}^{i+1}(n-1-t)^{\alpha - 1}(A_{i+1}^{j} - A_{i}^{j})dt \nonumber \\
		&=\frac{1}{\Gamma(\alpha + 1)}\sum_{i=0}^{n-2}(A_{i+1}^{j} - A_{i}^{j})((n-1-i)^{\alpha} - (n-2-i)^{\alpha})
	\end{eqnarray} 
	
	From here, regarding the second one, let $I = (i_{1}, \dots, i_{k}) \in W$ be a multi-index, it is enough to prove that ${}_{1}DS^{\alpha}(Y)_{0,1}^{I} = S^{\alpha}(Y)_{0,1}^{I}$. Nevertheless, as $Y$ is a linear path, we can directly conclude that: 
	\begin{eqnarray}
		&{}_{1}DS^{\alpha}(Y)_{0,1}^{I} = \frac{\prod_{j = 1}^{k}(A^{i_{j}}_{1} - A^{i_{j}}_{0})}{\Gamma(\alpha)\Gamma(1 + (k-1)\alpha)}(\frac{1^{\alpha}}{1})^{k}\beta_{\frac{1}{1}}((k-1)\alpha + 1, \alpha) = S^{\alpha}(Y)_{0,1}^{I}
	\end{eqnarray}
\end{proof}
\section{Application to machine learning}
Having introduced this new discrete fractional signature, we now develop a toy application in digit recognition to compare it with the classical one.

Indeed, using the well-known MNIST database (\citet{b7}), given $X^{m} = \{X_{i}^{m}\}_{i=0}^{783}$ a sequence of values corresponding an image, we embed a path from it as follows. First, we add to the value of each pixel its position, obtaining the new sequence $Y^{m} = \{Y_{i}^{m}\}_{i=0}^{783}$, where $Y_{i}^{m} = (x,y,X_{i}^{m})$, so that $i = 28x + y$ are given by the integer division. We then apply piecewise linear interpolation to this sequence in the interval $[0, 783]$ with equidistant points. 

From here, for both the new augmented training and test data sequences, we compute their discrete fractional signatures\footnote{Code used: \url{https://github.com/rubenjimlum/Fractional_signature}} and standardise them. Finally, we train and evaluate a machine learning gradient boosting model.

The results appear below. On the one hand, if we take the discrete fractional signature up to level 4, we can summarise the accuracy obtained for different values of $\alpha$ in the plot in Figure \ref{fig:image314}. Moreover, for the value $\alpha = 1.15$ (the best when we truncate at level 4), we compute the accuracy obtained truncating the discrete fractional signature at levels 4, 5, 6 and 7, and we compare the results with those of the original signature (table \ref{table}). 

In both cases, the discrete fractional signature beats the classical one.

\begin{minipage}{0.40\textwidth}
	\begin{table}[H]
			\centering
			\begin{tabular}{@{}lcc@{}}
				\hline
				& \multicolumn{2}{c}{Type of signature} \\
				\cline{2-3}
				Level &
				\multicolumn{1}{c}{$\alpha=1$} &
				\multicolumn{1}{c}{$\alpha=1.15$} \\
				\hline
				{Level 4} & 0.647   & 0.873 \\
				{Level 5} & 0.838   & 0.906\\
				{Level 6} & 0.881   & 0.924\\
				{Level 7} & 0.906   & 0.935\\
				\hline
			\end{tabular}
			\caption{Comparison of accuracies}
			\label{table}
		\end{table}
\end{minipage}
\hfill
\begin{minipage}{0.51\textwidth}
	\begin{figure}[H]
		\centering
		\includegraphics[scale=0.285]{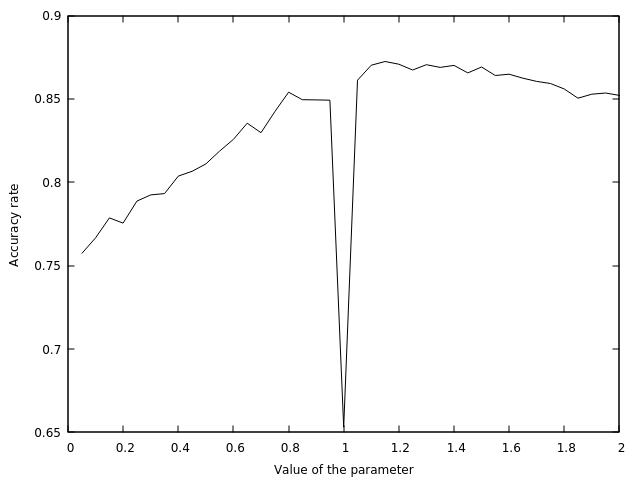}
		\caption{Accuracy as a function of $\alpha$}
		\label{fig:image314}
	\end{figure}
\end{minipage}
\section{Conclusions}
The theoretical results described in this paper encourage further study of the fractional signature. It remains to tackle the problem of considering the fractional signature in more general contexts and to study its uniqueness properties, because it may be able to characterize the path up to translations for certain values of $\alpha$. This is beyond the scope of this paper.

\end{document}